# Modeling Multi-Step Scientific Processes with Graph Transformer Networks

Amanda A. Volk[1], Robert W. Epps[2], Jeffrey G. Ethier[3], Luke A. Baldwin[3]

[1] *National Research Council Research Associate, Air Force Research Laboratory, Wright-Patterson Air Force Base, OH, 45433, United States*

[2] *Materials Science Center, National Renewable Energy Laboratory, Golden, CO 80401, United States*

[3] *Materials and Manufacturing Directorate, Air Force Research Laboratory, Wright-Patterson Air Force Base, OH 45433, United States*

## Abstract

This work presents the use of graph learning for the prediction of multi-step experimental outcomes for applications across experimental research, including material science, chemistry, and biology. The viability of geometric learning for regression tasks was benchmarked against a collection of linear models through a combination of simulated and real-world data training studies. First, a selection of five arbitrarily designed multi-step surrogate functions were developed to reflect various features commonly found within experimental processes. A graph transformer network outperformed all tested linear models in scenarios that featured hidden interactions between process steps and sequence dependent features, while retaining equivalent performance in sequence agnostic scenarios. Then, a similar comparison was applied to real-world literature data on algorithm guided colloidal atomic layer deposition. Using the complete reaction sequence as training data, the graph neural network outperformed all linear models in predicting the three spectral properties for most training set sizes. Further implementation of graph neural networks and geometric representation of scientific processes for the prediction of experiment outcomes could lead to algorithm driven navigation of higher dimension parameter spaces and efficient exploration of more dynamic systems.

**Introduction**

Multi-step experimental processes encompass most scientific procedures, ranging broadly from chemical syntheses to electronics fabrication to biomolecular processes. Modeling experiments from end-to-end remains highly challenging due to the large number of process parameters involved in even a simple multi-step protocol.[1] As a result, existing efforts to leverage machine learning in predicting the outcome of experiments have either relied on simplifying the experimental space into a few key parameters, as is typically found in self-driving laboratories and other works,[2–4] or focused on experimental processes with optimal substructure, as is found in retrosynthetic planning of organic syntheses.[5–8] Constraining parameter sets into smaller subsets of larger procedures imposes considerable limitations on the scope of conditions that can be algorithmically explored, and data sets collected over highly constrained spaces have little transferability between different studies, due to the resulting biases in sampling. Additionally, most experimental systems do not have optimal substructure that can be exploited as distinct segments of a larger process. Instead, procedures tend to feature steps that indeterminately influence the outcomes of later steps. These processes likely have hidden states and overlapping substructures that cannot be independently segmented. To capture this area of experimental processes, new models must be developed.

The recent artificial intelligence revolution has been largely fueled by the development of the transformer network and multi-headed attention mechanism, which is a dominant structure in modern large language and foundation models.[9] The transformer architecture offers significantly faster training times and higher predictability in many machine learning scenarios than prior methods, making them effective in natural language processing, among other applications.[10] In tandem with transformer networks, graph neural networks have been implemented across a range of modeling problems that can be represented by a graph structure, including network analyses and specific computer vision applications.[11–13] In many cases, the basic message passing neural network[14] has been augmented to provide a geometric parallel to popular linear neural network architectures, including the graph attention network,[15] graph convolutional network,[16] and graph transformer.[17] In the context of experimental systems, efforts have focused on the development of knowledge graphs for evaluating complete data environments,[18–20] molecular property prediction,[21,22] protein folding,[23,24] retrosynthetic planning, and chemical reactivity among others.[25–27] The application of graph neural networks in empirical science applications outside of these specific use cases is largely underexplored and warrants further study for potential points of viability.

In this work, we explore the application of geometric learning in predicting the outcome of multi-step scientific experimental procedures through directed acyclic graph representation. The viability of the geometric learning strategy was evaluated through predictive modeling of both simulated data sets and published real-world literature data. We first developed a series of surrogate ground truth functions meant to replicate potential behaviors embedded in experimental systems, and we compare the predictive performance of a graph transformer network to a series of linear models. In the initial simulation studies, we found that the graph models provided no considerable advantage in sequence agnostic surrogate functions. However, the graph neural networks considerably outperformed all linear models in every sequence dependent surrogate and every surrogate with overlapping substructure. Additionally, this difference grew further as the complexity of the surrogate space increased. Then, we evaluated data generated from recent work that studied colloidal atomic layer deposition in an automated droplet reactor, which presents a twenty-dimensional parameter space with three response features. While modeling the parameter set in full was unachievable both in the original publication and in our linear models, the graph neural network showed higher predictability for the first absorption peak wavelength, intensity, and peak to valley ratio. In this work, we demonstrate that geometric learning shows considerable promise in predicting the outcomes of the large, multi-step processes typically found in scientific procedures.

**Results**

Directed graph architectures can capture implicit sequential information that can be found typically in experimental procedures. To reflect the multi-step protocols as graph data structures, the process parameters were embedded into isolated nodes to represent each step, shown in Figure 1A. The nodes were then connected serially in a directed graph in order of their occurrence in the simulated procedure. Similarly, the linear data sets were constructed by appending all process parameters sequentially as a single input vector, shown in Figure 1B.

Data sets for the simulation studies were generated by randomly forming action sequences under specified constraints. Each step ($i$) features an action type ($A_i$), which is a one hot encoded categorical array with a specified maximum number of possible types ($N_{Types}$), and a set of numerical feature variables ($f_{i,j}$) of a specified length ($N_f$). The step sequence length was determined by randomly selecting an integer between 1 and the maximum action sequence length variable ($N_A$).

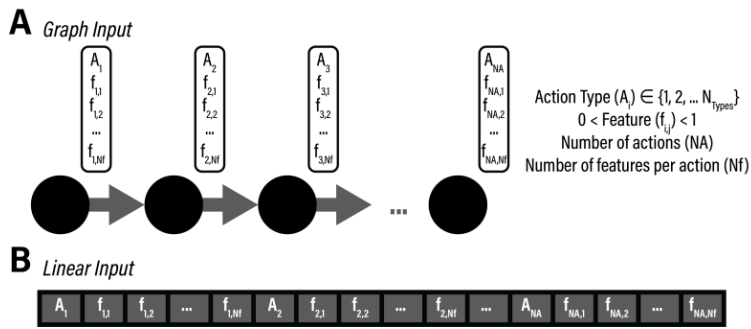

**Figure 1.** Illustration of input structure for (A) graph and (B) linear based models.

*Arbitrary Surrogates*

The four arbitrary surrogate ground truth functions were designed to reflect hidden features that could theoretically be found in real-world experimental systems. In these studies, the models could only access the final surrogate response value and could not access intermediate information. All surrogate function parameters were tuned so that the output response distribution would range approximately from zero to one to enable more direct comparisons between surrogates. By evaluating model performance on these surrogates, we can roughly approximate the potential of various model architectures in their application in the real world. First, the sequence agnostic surrogate, shown in Figure 2A, represents process steps that do not interact spatially between each other. Every step equally and unbiasedly contributes to the ultimate surrogate response. This type of interaction is likely to be found in more robust and less reactive processes, such as stable equilibrium systems and inert materials mixing. The sequence agnostic surrogate is expected to perform well with the same model architectures that effectively navigate black box systems. Next, a sequence dependent surrogate, shown in Figure 2B, was developed to evaluate protocols that exhibit a time correlated behavior in process steps. This type of interaction in theory encompasses most experimental procedures, such as chemical precursors that form intermediate complexes based on the order of material addition, handling of environmental exposures in a multi-step process, or any electronics fabrication procedure.

The final two arbitrary surrogate functions are designed to capture overlapping substructure mechanisms. The first function illustrated in Figure 2C, called the critical sequence surrogate, is a sequence dependent function with a changing power term at step $i$ ($p_i$). The surrogate sequentially steps through the input features at each step uniformly unless the action type category features begin to follow a specified sequence. When the current step categorical falls within the required sequence, the power term is increased, resulting in an increased influence of later steps in the protocol. This type of interaction is expected to occur in processes that feature a critical selection of steps that disproportionately influence the reaction outcome, such as the deposition of the active layer in solar cells or the growth phase of a material after preparing precursors. Finally, we evaluated a delayed interaction surrogate, shown in Figure 2D. The delayed interaction surrogate is sequence agnostic when the step type features are different, but when the current step category matches that of the prior steps, the numerical features of all matching prior steps are merged into a single numeric parameter. This type of interaction results in unevenly distributed interaction effects that are primarily marked by the categorical type. This type of delayed interaction could be found in experimental features such as environmental controls and contamination or unexpected chemical reactivity.

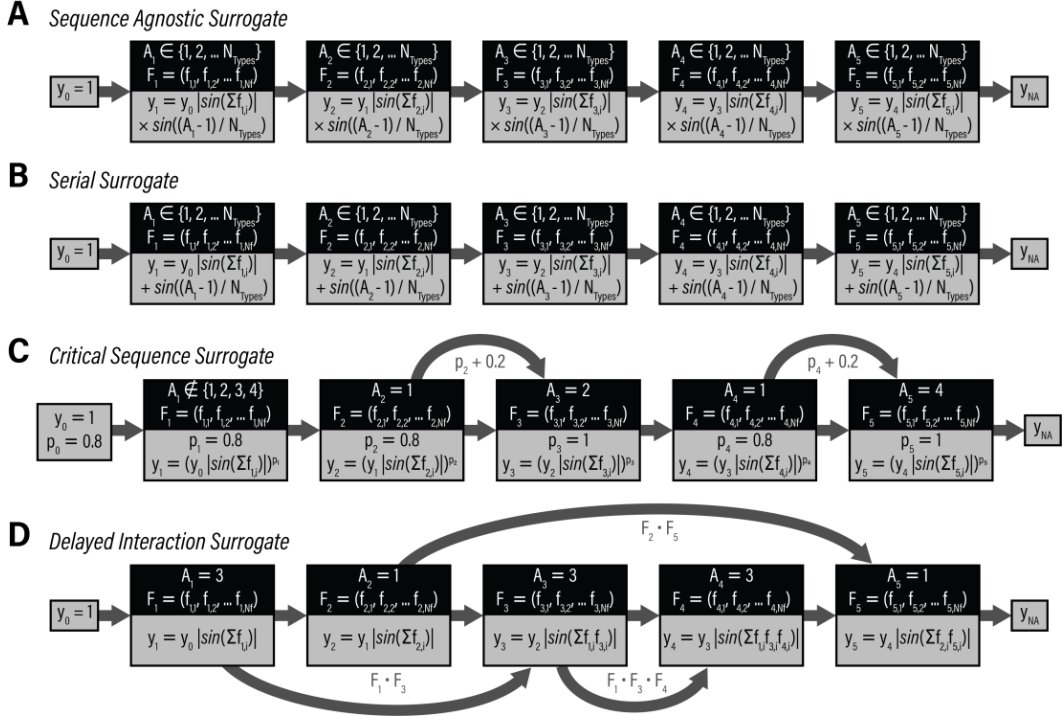

**Figure 2.** Illustrations of the arbitrary surrogates. Sample step sequences for the (A) sequence agnostic surrogate, (B) sequence dependent surrogate, (C) critical sequence cluster surrogate, and (D) delayed interactions surrogate.

Using the step sequence generation system, data sets were built using each of the four surrogates and used to train a set of models – a gradient boosted decision tree regressor (GBDT), a gaussian process regressor (GP), a support vector machine regressor (SVM), a feed forward neural network (LNN), and a graph transformer network (GTN). The models were each trained on data set sizes ranging from 50 to 6,400 sequences, each randomly generated before training. Each condition set was replicated ten times.

In the sequence agnostic system, shown in Figure 3A, the GTN, LNN, and GP achieved mean squared errors (MSEs) of less than 0.001 within 800 samples, and the SVM and GBDT each reached below 0.002 after 6,400 samples. Among the linear models the LNN and GPR performed the best with the SVM and GBDT appearing to begin convergence onto a suboptimal model after roughly 1600 samples. More surprisingly, the GTN had slightly improved performance over the LNN and GPR in this space for most training set sizes. This advantage could be attributed to the clustering of features within individual steps of the surrogate. The innately clustered structure of feature data embedded into individual nodes could generate structural similarities to the surrogate that the model is trying to predict. Alternatively, further hyperparameter tuning or exploration of alternative linear model architectures could lead to more competitive performance. While the GTN exhibits a notably longer training time than either linear model, it would serve as suitable predictor for similar black box systems without sequence features. The results of the sequence dependent surrogate, shown in Figure 3B, show a clearer distinction between the GTN and linear models.

The GTN predictions had a lower MSE than all models for every training set size above 100 sequences and less than a third of the MSE of the best linear models for training set sizes of 1,600, 3,200, and 6,400 sequences. This significant improvement over linear models suggests that the sequence data inherent to directed graph structures could play a notable role in forming predictive correlations. This property would likely contribute to higher predictability in real-world processes, given sufficient data.

Among the arbitrary surrogates, the critical point surrogate, shown in Figure 3C, generated the most significant difference between the geometric and linear models. The GTN had the lowest MSE among the test models for all training set sizes above 400 sequences. Additionally, the MSE of the GTN at 6,400 sequences is less than one tenth of the MSE of the next highest performing model – the LNN – indicating a substantial improvement in predictability. The combination of clustered substructure dependencies and sequence-based response found in the critical sequence surrogate reveals a significant strength of GTN predictions. It should be noted that this improvement is most likely not attributable to an excess of padding in the linear data architecture. Shown in supporting information Figure S.1, when the sequence of steps is set to a fixed length instead of the randomly selected length used in the rest of the arbitrary surrogate studies, the model behaviors on the critical point surrogate are very similar. The GTN performs significantly better than all the linear models, and some linear models suffer a slight reduction in performance likely due to a large percentage of the test data set comprising of larger, more complex sequences.

Similarly, the results from the delayed interaction surrogate, shown in Figure 3D, showed that GTNs more effectively predicted the response of overlapping substructure units than linear models. However, the improvement of GTNs on the delayed interaction surrogate are substantially smaller than those of the critical point surrogate. While the GTN outperformed all linear models for training set sizes above 400 samples, the MSE never reached above a 50% improvement over the next best model. This loss in advantage could be attributed to several factors. First, the delayed interaction surrogate is sequence agnostic for many process steps, which would reduce the improvement gain from the directed graph architecture. Second, in the cases where the protocol steps match prior step types and the numerical features are sequence dependent, the relevant steps exhibit considerable delay and disconnect as an interactive group. The size and positioning of the correlated steps will be randomly distributed, making them more challenging to group together effectively. Finally, the nature of the selected surrogate function, multiplicative, would cause the sequence dependent features to appear less prominent than that of the critical sequence surrogate, which is exponential. If this influence were adjusted to be larger relative to the average response behavior, then it is likely that the GTNs would outperform the linear models with a greater magnitude. The observation of any improvement over the purely sequence agnostic surrogate suggests that the GTN can capture more subtle and overlapping substructures than the tested linear models.

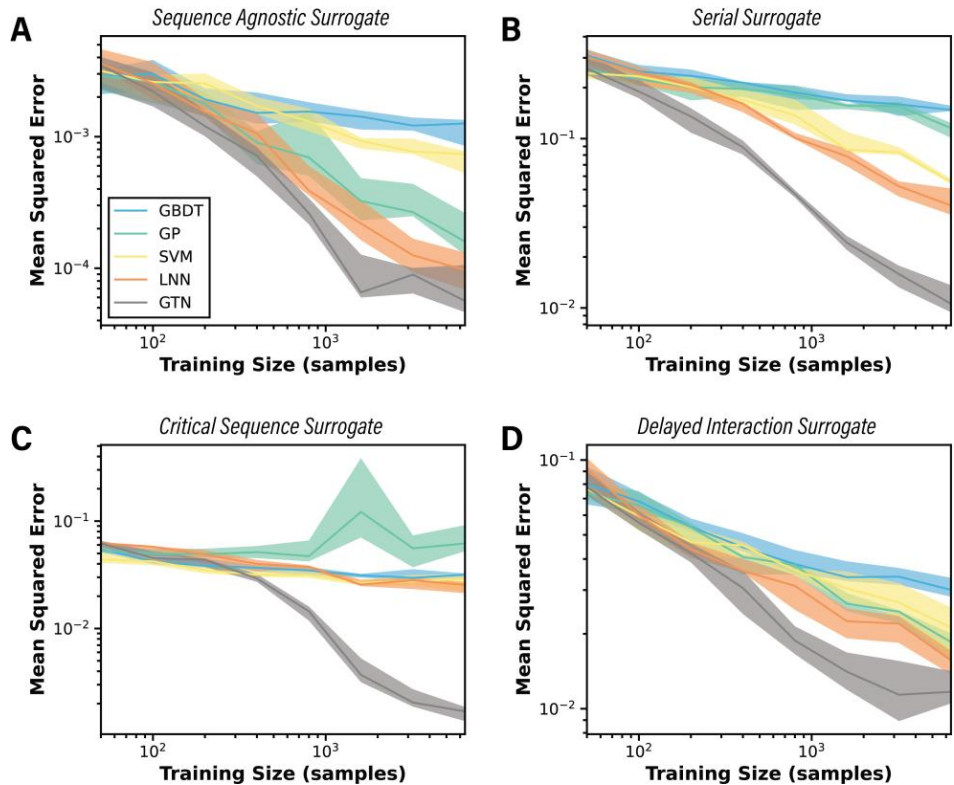

**Figure 3.** Arbitrary surrogate training results. The median mean squared error across ten replicates as a function of the training set size for the (A) sequence agnostic surrogate, (B) sequence dependent surrogate, (C) critical point surrogate, and (D) delayed effect surrogate using a gradient boosted decision tree regressor [GBDT], a gaussian process regressor [GP], a support vector machine regressor [SVM], a linear neural network [LNN], and a graph transformer convolutional network [GTN]. The solid line represents the median of the replicates, and the shaded region represents the inner quartile range.

Finally, a fifth surrogate, the sum effect surrogate, was built using the combined sum of the four surrogates to evaluate the effects of model complexity and the ability to simultaneously model the various surrogate features. Within this sum effect surrogate, the GTN excels in more complex variants of the surrogate functions. Shown in Figure 4, the sum effect surrogate simulations were replicated at four different levels of surrogate complexity with the same training models. With a simpler surrogate, i.e. with fewer parameters, the linear models perform suitably, and the GP even outperforms the GTN at higher data set sizes. This advantage disappears for the moderate, high, and very high complexity surrogates. After 6,400 training sequences on the very high complexity surrogate, the GTN achieves half of the median mean squared error of the next best model. Similar trends can be observed with the critical point, sequence dependent, and the delayed interactions surrogates, show in supporting information Figures S.2 to S.5.

The tendency of the GTN to perform exceptionally on high complexity surrogates and with larger data set sizes – greater than 1,000 – presents an important opportunity in experimental research.

To-date, modeling experimental outcomes through process parameters has typically relied on some simplification of the parameter set, through explicit segmentation of process steps, constraining parameter spaces, or the introduction of physics-based features. GTNs and likely other transformer-based architectures could potentially be coupled with increased and more diverse collection and reporting of process information and outcomes. Like the role of transformer networks in preceding modern multifunctional foundation models, GTNs and the associated graphical representation of complex multi-step experimental processes could lead to the formation of experimental research foundation models and generalized scientific procedure generation and prediction algorithms.

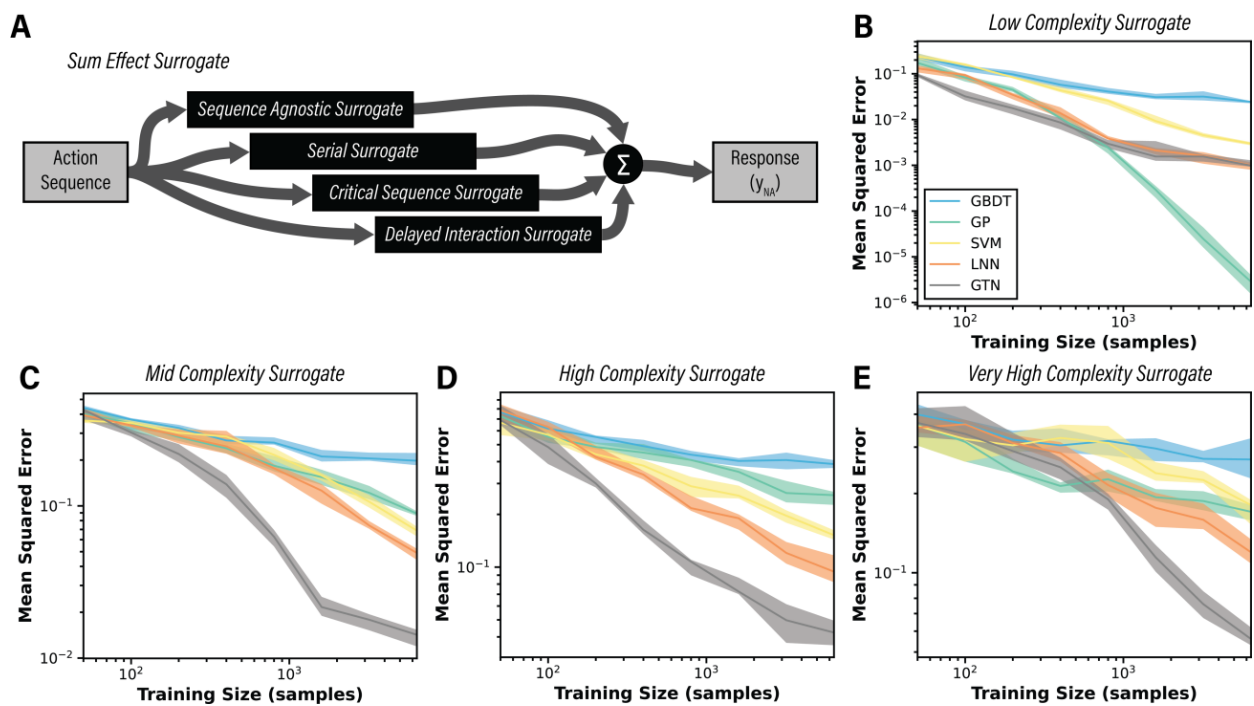

**Figure 4.** All feature sum surrogate simulations under different complexity parameters. (A) Illustration of the sum effect surrogate design. Mean squared error as a function of the training set size for the sum effect surrogate with (B) 3 maximum steps, 5 step types, and 2 numerical features per step, (C) 5 maximum steps, 7 step types, and 3 numerical features per step, (D) 7 maximum steps, 9 step types, and 5 numerical features per step, and (E) 9 maximum steps, 11 step types, and 7 numerical features per step using a gradient boosted decision tree regressor [GBDT], a gaussian process regressor [GP], a support vector machine regressor [SVM], a linear neural network [LNN], and a graph transformer convolutional network [GTN]. The solid line represents the median of the replicates, and the shaded region represents the inner quartile range.

*Colloidal Atomic Layer Deposition Data Set*

Finally, the efficacy of GTNs in modeling complex multi-step processes was evaluated from a real-world data set taken from literature. In recent work, the platform *AlphaFlow* was leveraged to navigate the multistep process of colloidal atomic later deposition on reactive microdroplets using reinforcement learning.[28] In the work, precursors were added in an algorithmically determined sequence and measurements were taken after each addition with the aim of artificially reconstructing the complex process of colloidal atomic layer deposition. As the size of this parameter space reached 20 dimensions, typical linear models were unable to provide sufficient predictability when evaluating the process from end-to-end.

In this work, we reevaluate the data generated in this sequence selection study but with the inclusion of GTNs. The same collection of models applied previously in this manuscript were adapted to the *AlphaFlow* data set to include intermediate sampling data. For the linear models, the input array was built by sequentially adding the precursor selection value and the previous measurement value for each injection step, then the remainder of the array was padded to match the maximum number of possible features. Similarly, the graph data set was formatted so that each node contained the precursor selection and the prior measurement value, then the nodes were connected in a directed graph format to correspond to the sequence of injections. The full data set, comprising of approximately 120 full droplet injection sequences, was divided by droplet into training sets of a specified length and testing sets of 25 droplets. The three response parameters were adjusted with min-max scaling for model training and mean squared error comparisons. Each training conditions were conducted with randomized training and testing set generation for 20 replicates.

Shown in Figure 6, the GTNs were able to consistently outperform all linear methods in predicting the first absorption peak wavelength, absorption peak to valley ratio, and first absorption peak intensity after 20 training droplets. The GBDT, GP, and SVM each showed little improvement in predictive capabilities with the addition of more data, seeming to plateau throughout the full training set size range. The LNN performed favorably in predicting the first absorption peak wavelength, but failed to improve over the training set sizes for the peak to valley ratio or peak intensity.

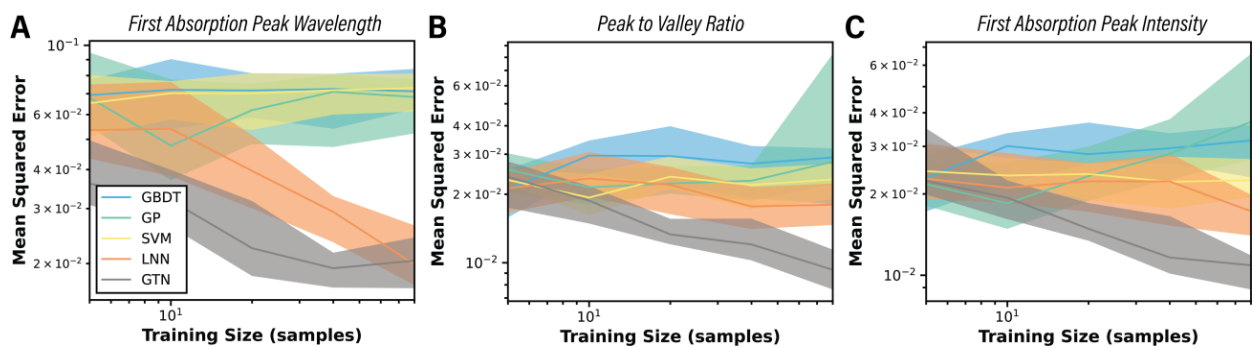

**Figure 6.** Modeling with the *AlphaFlow* sequence selection campaign. The mean squared error of the test set model prediction for the (A) first absorption peak wavelength, (B) absorption peak to valley ratio, and (C) absorption peak intensity with intermediate data as a function of the number of training droplets using a gradient boosted decision tree regressor [GBDT], a gaussian process regressor [GP], a support vector machine regressor [SVM], a linear neural network [LNN], and a graph transformer convolutional network [GTN]. The solid line represents the median of the replicates, and the shaded region represents the inner quartile range.

**Conclusions**

This work extends the application of graph learning and graph transformers into the space of multi-step experimental processes. For the first time, we have demonstrated that representation of a multi-step procedure as a single graph structure can enable more effective process outcome modeling than competitive linear modeling strategies. Through simulations on arbitrary surrogate functions, we identified case scenarios where the use of GTNs provided significantly improved predictability. Specifically, GTNs outperformed the tested linear models in settings with hidden and overlapping substructure, spatially clustered step interactions, and sequence dependent process interactions. Additionally, we found that there was no considerable penalty in scenarios without these sequence dependent interactions, and the GTN performed equivalently to linear models. The one scenario where GTNs performed worse than the best tested linear models was in low complexity surrogate spaces, where GPs excelled at high precision predictions with sufficiently large training set sizes. We also showed that GTNs exhibit viability in real-world experimental scenarios. We demonstrated the viability of GTNs in predicting the outcomes of real experimental data, taken from recent literature on identifying colloidal atomic layer deposition pathways. GTNs more accurately predicted all three spectral properties than all linear models for most tested training set sizes.

The implementation of geometric learning in this work is constrained to graph-level prediction of a single experimental response on a serially connected graph architecture. The capabilities and versatility of geometric learning techniques extend far beyond this specific use-case, and exploration of branching directed graph systems, among other possibilities, could provide even greater advantages than those shown in this work. There is also a high probability that more efficient implementations of geometric learning in experimental systems are available. The topic warrants further exploration and implementation. Graphical representation of experimental processes could lead to more effective prediction of experimental outcomes in dynamically defined experimental spaces. This strategy could generate an improved understanding and more efficient algorithm guided exploration of large, unconstrained experimental spaces, resulting in experimental algorithms that more closely reflect the versatility of research in the real world.

**Methods**

*Modeling*

The gradient boosting decision tree regressor, support vector machine regressor, and gaussian process regressor models were from the Scikit-Learn library.[29] The linear neural network was built using the Pytorch library[30,31] and was comprised of four descending size linear layers with ReLU activation functions and an Adam optimizer. The graph transformer network was built using the Pytorch-Geometric library[32] and comprised of four constant size multiheaded attention graph transformer layers,[17] global mean and global max pooling, and four descending size linear layers with ReLU activation functions and an Adam optimizer.

*Arbitrary Surrogate Generation*

For the arbitrary surrogate simulations, data sets were constructed using a random action sequence generator function. The generator would create randomized action sequences within the constraints specified by the target complexity value. The generator parameters consisted of the maximum number of action steps, the total number of possible action types, and the total number of continuous features per action. For each simulated experiment, the generator would randomly select a number of action steps between one and the maximum action steps parameter, then each step would be filled in with the randomly selected one-hot encoded action step type category followed by the randomly generated continuous numerical features. The low complexity generator used 3 maximum steps, 5 step types, and 2 numerical features per step. The moderate complexity generator used 5 maximum steps, 7 step types, and 3 numerical features per step. The high complexity generator used 7 maximum steps, 9 step types, and 5 numerical features per step. Finally, the very high complexity generator used 9 maximum steps, 11 step types, and 7 numerical features per step.

*Colloidal Atomic Layer Deposition (cALD) Training Design*

The input data set for both the graph and linear models were structured similarly for the cALD studies as they were for the arbitrary surrogate studies, except for the inclusion of intermediate measurements. In these studies, each action step was represented by the one-hot encoded precursor selection along with the previous step spectra measurement. In the graph design, this action step would be embedded within a single node of the input graph, and the final spectral measurement was represented normally through the graph level prediction. In the linear design each action step was appended to a single input array with padding for uneven injection numbers between droplets.

*Supporting Information*

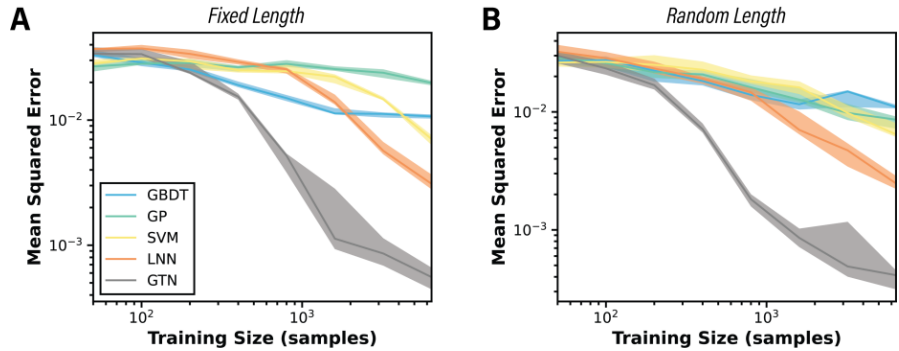

**Fig S.1.** Effect of fixed versus variable length action sequences. The median mean squared error across ten replicates as a function of the training set size for the critical point surrogate with (A) a fixed action length of 5 steps and (B) a randomly selected action length between 1 and 5 steps using a gradient boosted decision tree regressor [GBDT], a gaussian process regressor [GP], a support vector machine regressor [SVM], a linear neural network [LNN], and a graph transformer convolutional network [GTN]. The solid line represents the median of the replicates, and the shaded region represents the inner quartile range.

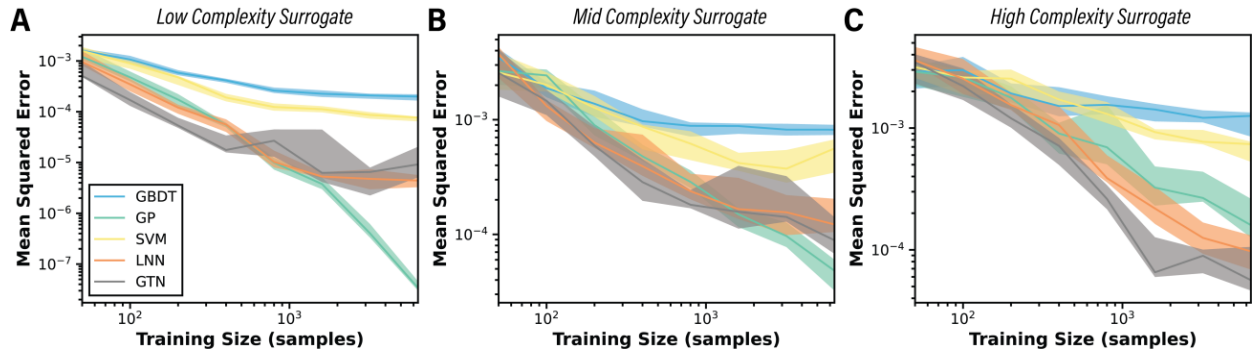

**Fig S.2.** Sequence agnostic surrogate simulation results at different complexities. Mean squared error as a function of the training set size for the sequence agnostic surrogate with (A) 3 maximum steps, 5 step types, and 2 numerical features per step, (B) 5 maximum steps, 7 step types, and 3 numerical features per step, and (C) 7 maximum steps, 9 step types, and 5 numerical features per step using a gradient boosted decision tree regressor [GBDT], a gaussian process regressor [GP], a support vector machine regressor [SVM], a linear neural network [LNN], and a graph transformer convolutional network [GTN]. The solid line represents the median of the replicates, and the shaded region represents the inner quartile range.

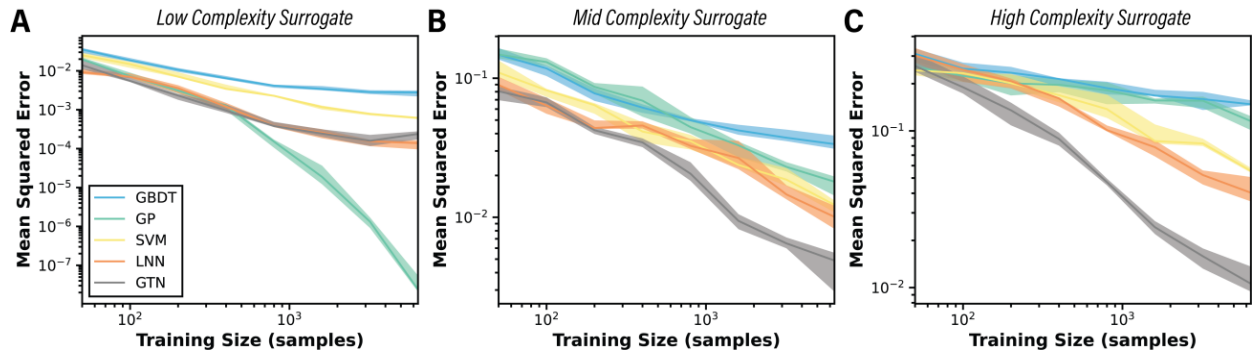

**Fig S.3.** Sequence dependent surrogate simulation results at different complexities. Mean squared error as a function of the training set size for the sequence dependent surrogate with (A) 3 maximum steps, 5 step types, and 2 numerical features per step, (B) 5 maximum steps, 7 step types, and 3 numerical features per step, and (C) 7 maximum steps, 9 step types, and 5 numerical features per step using a gradient boosted decision tree regressor [GBDT], a gaussian process regressor [GP], a support vector machine regressor [SVM], a linear neural network [LNN], and a graph transformer convolutional network [GTN]. The solid line represents the median of the replicates, and the shaded region represents the inner quartile range.

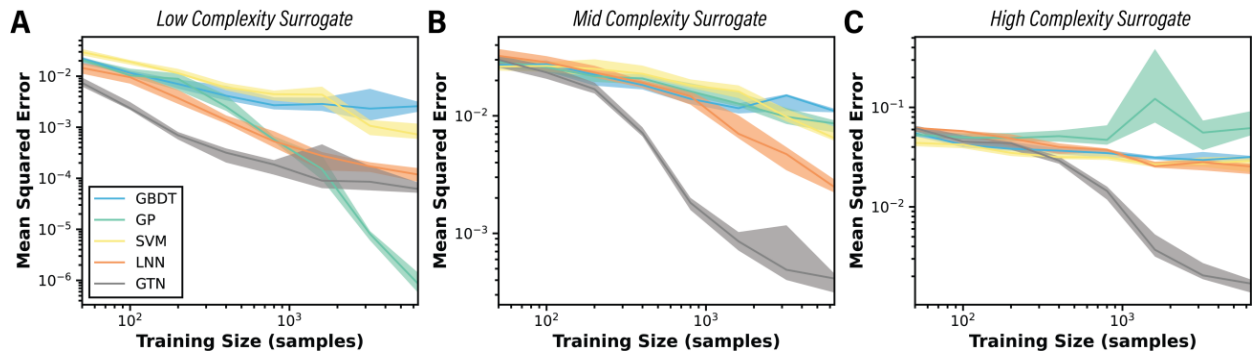

**Fig S.4.** Critical sequence surrogate simulation results at different complexities. Mean squared error as a function of the training set size for the critical sequence surrogate with (A) 3 maximum steps, 5 step types, and 2 numerical features per step, (B) 5 maximum steps, 7 step types, and 3 numerical features per step, and (C) 7 maximum steps, 9 step types, and 5 numerical features per step using a gradient boosted decision tree regressor [GBDT], a gaussian process regressor [GP], a support vector machine regressor [SVM], a linear neural network [LNN], and a graph transformer convolutional network [GTN]. The solid line represents the median of the replicates, and the shaded region represents the inner quartile range.

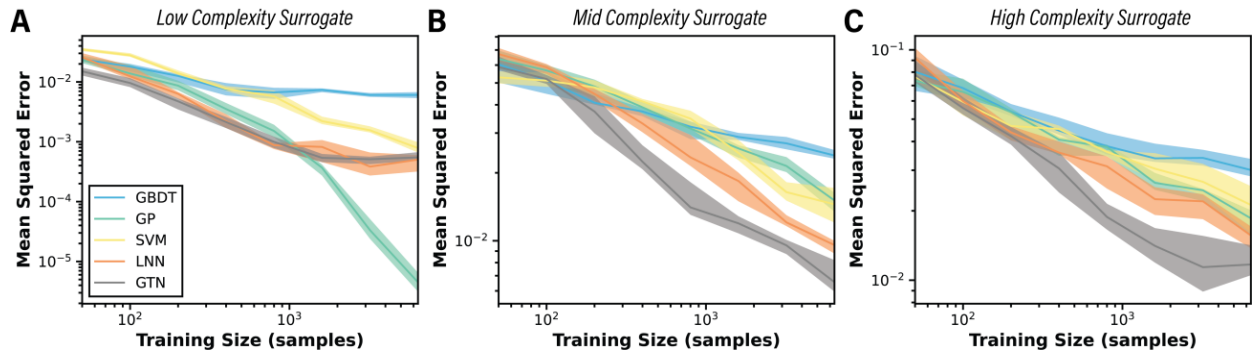

**Fig S.5.** Delayed interaction surrogate simulation results at different complexities. Mean squared error as a function of the training set size for the delayed interaction surrogate with (A) 3 maximum steps, 5 step types, and 2 numerical features per step, (B) 5 maximum steps, 7 step types, and 3 numerical features per step, and (C) 7 maximum steps, 9 step types, and 5 numerical features per step using a gradient boosted decision tree regressor [GBDT], a gaussian process regressor [GP], a support vector machine regressor [SVM], a linear neural network [LNN], and a graph transformer convolutional network [GTN]. The solid line represents the median of the replicates, and the shaded region represents the inner quartile range.